\documentclass[12pt,english]{article}
\pdfoutput=1
\usepackage{amsmath,amsfonts,amsthm,amssymb}
\usepackage{array}
\usepackage{setspace}
\usepackage{Tabbing}
\usepackage{graphicx,float}
\usepackage{slashbox}
\usepackage{booktabs}
\usepackage[authoryear]{natbib}
\usepackage[colorlinks,citecolor=blue,
    linkcolor=black]{hyperref}
\hypersetup{bookmarksnumbered}
\usepackage{geometry}
\usepackage{multirow}
\usepackage{arydshln}

\allowdisplaybreaks[3]
\newcommand{\ra}[1]{\renewcommand{\arraystretch}{#1}}

\DeclareMathOperator*{\argmin}{arg\ min}
\DeclareMathOperator*{\argmax}{arg\ max}

\usepackage[width=\textwidth ,font={small},labelfont={color=blue,bf,up},
labelsep=endash, format=plain, labelsep=period, 
textfont=it]{caption}

\newtheorem{Lemma}{Lemma}
\numberwithin{equation}{section}

\floatstyle{ruled}
\newfloat{algorithm}{tbp}{loa}
\providecommand{\algorithmname}{Algorithm}
\floatname{algorithm}{\protect\algorithmname}

\newcommand{\bs}{\boldsymbol}

\makeatletter
\def\Lemma@space@setup{\Lemma@preskip=0pt
\Lemma@postskip=0pt}
\makeatother

\usepackage{multicol}
\begin{document}

\title{\LARGE \bf Sparse Distance Weighted Discrimination}

 \author{\sc{Boxiang Wang\thanks{School of Statistics, University of Minnesota.}} and \sc{Hui Zou} \thanks{Corresponding author, zouxx019@umn.edu. School of Statistics, University of Minnesota.}\\
 \date{First Version: Jun 11, 2014\\
 Second Version: Jan 04, 2015}
}
\maketitle
\begin{abstract}
Distance weighted discrimination (DWD) was originally proposed to handle the data piling issue in the support vector machine. In this paper, we consider the sparse penalized DWD for high-dimensional classification.
The state-of-the-art algorithm for solving the standard DWD is based on second-order cone programming, however such an algorithm does not work well for the sparse penalized DWD with high-dimensional data.
In order to overcome the challenging computation difficulty, we develop a very efficient algorithm to compute the solution path of the sparse DWD at a given fine grid of regularization parameters. We implement the algorithm 
in a publicly available R package \texttt{sdwd}. We conduct extensive numerical experiments to demonstrate the computational efficiency and classification performance of our method. 
\end{abstract}
\newpage

\section{Introduction}
The support vector machine (SVM) \citep{Vapnik1995} is a widely used modern classification method. In the standard binary classification problem, training dataset consists of $n$ pairs, $\{(\mathbf{x}_i, y_i)\}_{i=1}^n$, where $\mathbf{x}_i\in\mathbb{R}^p$ and $y_i \in \{-1, 1\}$. The linear SVM seeks a hyperplane $\{\mathbf{x}: \beta_0 + \mathbf{x}^T\bs\beta = 0 \}$
which maximizes the smallest margin of all data points:
\begin{align}
\argmax_{\beta_0, \bs{\beta}}\;\;\;\;& \min_i d_i,\notag\\
\text{subject to }\;\;\;&  d_i = y_i (\beta_0 + \mathbf{x}_i^T\bs{\beta})+\eta_i \ge 0, \ \forall i,\notag\\
&\eta_i \ge 0, \ \forall i, \ \sum_{i=1}^n \eta_i \le c, \ ||\bs{\beta}||_2^2 = 1,\label{eq:SVM1}
\end{align}
where $d_i$ is defined as the \textit{margin} of the $i$th data point, $\eta_i$'s are slack variables introduced to ensure all margins non-negative, and $c>0$ is a tuning parameter controlling the overlap. By using a kernel trick, the SVM can also produce nonlinear decision boundaries by fitting an optimal separating hyperplane in the extended kernel feature space.  The readers are referred to \cite{HastieEtAl2009} for a more detailed explanation of the SVM.  

\cite{MarronEtAl2007} noticed that when the SVM is applied on some data with $n < p$, many data points lie on two hyperplanes parallel to the decision boundary. \cite{MarronEtAl2007} referred to this phenomenon as \textit{data pilling} and claimed that the data pilling can ``affect the generalization performance of SVM". To overcome this issue, \cite{MarronEtAl2007} proposed a new method called the distance weighted discrimination (DWD), which finds a separating hyperplane minimizing the sum of the inverse margins of all data points:
\begin{align}
\argmin_{\beta_0, \bs{\beta}}\;\;\;\;& \sum_i {1}/{d_i},\notag\\
\text{subject to }\;\;\;&  d_i = y_i (\beta_0 + \mathbf{x}_i^T\bs{\beta})+
\eta_i \ge 0, \ \forall i,\notag\\
&\eta_i \ge 0, \ \forall i, \ \sum_i \eta_i \le c, \ ||\bs{\beta}||_2^2 = 1.
\label{eq:DWD1}
\end{align}
The initial version of \cite{MarronEtAl2007} also mentioned the sum of the inverse margins $\sum_i 1/d_i$ could be also replaced by $\sum_i 1/d_i^q$, the $q$th power of the inverse margins, and this generalized version was used as the definition of the DWD in \cite{HallEtAl2005}. \cite{MarronEtAl2007} asserted the DWD can avoid the data piling and thereby improve the generalizability. One example [see the group 2 of Figure 3 in \cite{MarronEtAl2007}] shows that the DWD has about 5\% prediction error whereas the SVM does 15\%.  Enhancement of the DWD over the SVM can also be exemplified in \cite{HallEtAl2005} through a novel geometric view. As for the computation of the DWD, \cite{MarronEtAl2007} observed that the DWD is an application of the second-order cone programming and thus can be solved by the primal-dual interior-point methods. The algorithm has been implemented in both Matlab code \url{http://www.unc.edu/~marron/marron_software.html} and an R package \texttt{DWD} \citep{HuangEtAl2012}. 

In this paper we focus on classification with high-dimensional data where the number of covariates is much larger than the sample size. The standard SVM and DWD are not suitable tools for high-dimensional classification for two reasons. First, based on the scientific hypothesis that only a few important variables affect the outcome, a good classifier for high-dimensional classification should have the ability to select important variables and discard irrelevant ones. However, the standard SVM and DWD use all variables and do not conduct variable selection. Second, because these two classifiers use all variables, they may have very poor classification performance. As explained in \cite{FanFan2008}, the bad performance is caused by the error accumulation when estimating too many noise variables in the classifier. Owing to these two considerations, sparse classifiers are generally preferred for high-dimensional classification. In the literature, some penalties have been applied to the SVM to produce sparse SVMs such as the $\ell_1$ SVM \citep{BradleyMangasarian1998, ZhuEtAl2004}, the SCAD SVM \citep{Zhang2006}, and the elastic-net penalized SVM \citep{WangEtAl2006}.

\begin{figure}
\centering
\begin{minipage}[!h]{.49\textwidth}
\centering
    \includegraphics[width=\textwidth]{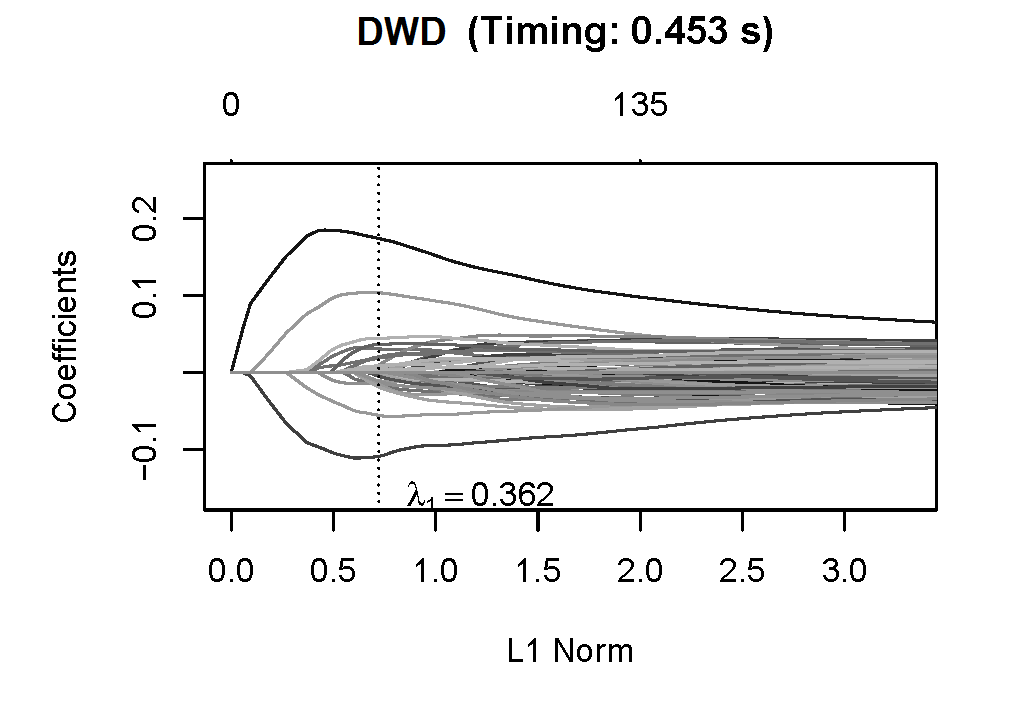}
\end{minipage}
\begin{minipage}[!h]{.49\textwidth}
\centering
    \includegraphics[width=\textwidth]{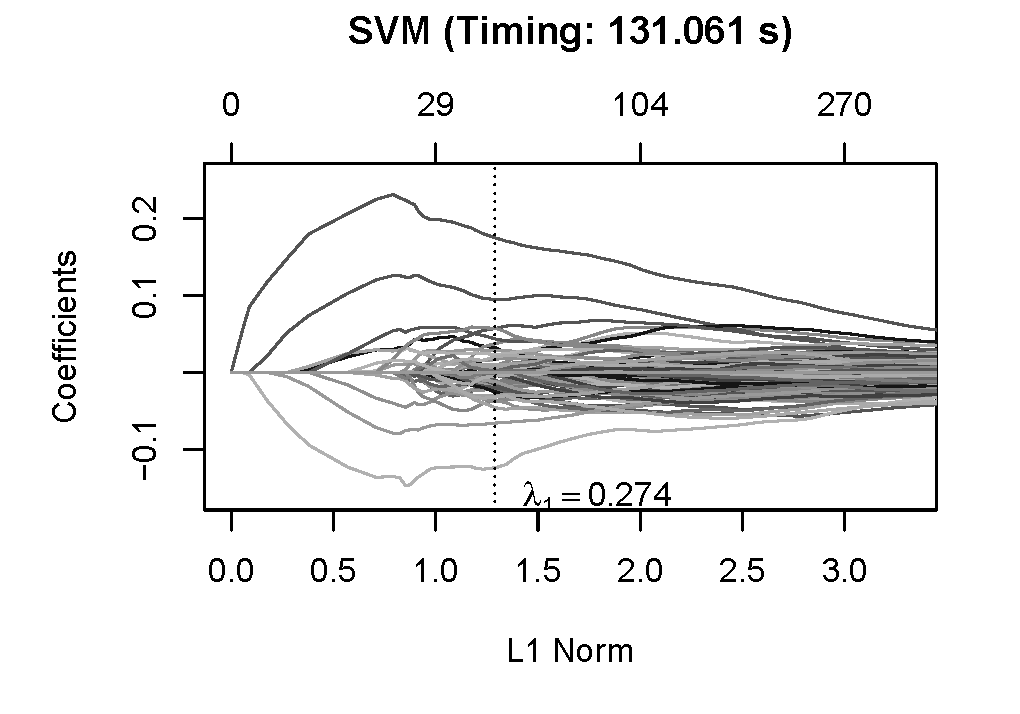}
\end{minipage}
 \caption{The solution paths for the prostate data ($n=102$, $p=6033$) using the elastic-net DWD and the elastic-net SVM. In every method, $\lambda_2$ is fixed to be 1. The dashed vertical lines indicate the $\lambda_1$ selected by the five-folder cross validation. Both timings are averaged over 10 runs. }
 \label{fig:solnpaths}
\end{figure}

In this work we consider sparse penalized DWD for high dimensional classification. The standard DWD uses the $\ell_2$ penalty and can be solved by the second-order cone programming. However, the sparse DWD is computationally more challenging and requires a different computing algorithm. To cope with the computational challenges associated with the sparse penalty and high-dimensionality, we derive an efficient algorithm to solve the sparse DWD by combining majorization-minimization principle and coordinate-descent. We have implemented the algorithm in an R package \texttt{sdwd}. To give a quick demonstration here, we use the prostate cancer data [\cite{SinghEtAl2002}, 102 observations and 6033 genes] as an example. The left panel of Figure~\ref{fig:solnpaths} depicts the solution paths of the elastic-net penalized DWD, and \texttt{sdwd} only took 0.453 second to compute the whole solution path. As comparison, we also used the code in \cite{WangEtAl2006} to compute the solution path of the elastic-net penalized SVM. 
We observed that the timing of the sparse SVM was about 290 times larger than that of the sparse DWD.

\section{Sparse DWD}
\label{sec:SPF}
In this section we present several sparse penalized DWDs. Our formulation follows the $\ell_1$ SVM \citep{ZhuEtAl2004}. Thus, we first review the derivation process of the $\ell_1$ SVM. 
The standard SVM \eqref{eq:SVM1} is often rephrased as the following quadratic programming problem \citep{HastieEtAl2009}:
\begin{equation}
\begin{aligned}
\argmin_{\beta_0, \bs{\beta}}\;\;\;\;& ||\bs\beta||^2_2
\label{eq:l2SVM}\\
\text{subject to }\;\;\;&  y_i (\beta_0 + \mathbf{x}_i^T\bs{\beta})+\eta_i \ge 1, \ \forall i,\notag\\
&\eta_i \ge 0, \ \forall i, \ \sum_{i=1}^n \eta_i \le c.\notag
\end{aligned}
\end{equation}
Moreover, the above constrained minimization problem has an equivalent \textit{loss+penalty} formulation \citep{HastieEtAl2009}:
$$
\argmin_{\beta_0, \bs{\beta}}\dfrac{1}{n}\sum_{i=1}^n\left[1-y_i (\beta_0 + \mathbf{x}_i^T\bs{\beta})\right]_{+} + \dfrac{\lambda_2}{2}||\bs{\beta}||^2_2. \label{eq:l2SVMloss}
$$
The loss function $[1-t]_{+}=\max(1-t,\ 0)$ is the so-called hinge loss in the literature. 
For the high-dimensional setting, the standard SVM uses all variables because of the $\ell_2$ norm penalty used therein. As a result, its performance can be very poor. 
\cite{ZhuEtAl2004} proposed the $\ell_1$-norm SVM to fix this issue:
$$
\argmin_{\beta_0, \bs{\beta}}\dfrac{1}{n}\sum_{i=1}^n\left[1-y_i (\beta_0 + \mathbf{x}_i^T\bs{\beta})\right]_{+} + \lambda_1||\bs{\beta}||_1. \label{eq:l1SVMloss}
$$

Similarly, we can propose the $\ell_1$ penalized DWD. It has been shown that the standard DWD also has a \textit{loss+penalty} formulation  \citep{LiuEtAl2011}:
$$
\argmin_{\beta_0, \bs{\beta}}\dfrac{1}{n}\sum_{i=1}^n V\left(y_i (\beta_0 + \mathbf{x}_i^T\bs{\beta})\right) +\dfrac{\lambda_2}{2}||\bs{\beta}||^2_2, 
$$
where the loss function is given by
$$
V(u)=
\begin{cases}
1-u, \; &\text{ if } u \le  1/2,\\
      1/(4u), \; &\text{ if } u > 1/2.
    \end{cases}
\label{eq:HL19}
$$
Similar to the $\ell_1$ SVM, we replace the $\ell_2$ norm penalty with the $\ell_1$ norm penalty in order to achieve sparsity in the DWD classifier. Hence, the $\ell_1$ DWD is defined by
\begin{equation}
\left(\hat\beta_0(\mathrm{lasso}),\hat{\bs{\beta}}(\mathrm{lasso})\right)=\argmin_{\beta_0, \bs{\beta}}\dfrac{1}{n}\sum_{i=1}^n V\left(y_i (\beta_0 + \mathbf{x}_i^T\bs{\beta})\right) +{\lambda_1}||\bs{\beta}||_1.
\end{equation}
The lasso penalized DWD classification rule is $\textrm{Sign}(\hat\beta_0(\mathrm{lasso}) + \mathbf{x}^T\hat{\bs{\beta}}(\mathrm{lasso}))$.

Besides the $\ell_1$ norm penalty, we also consider the elastic-net penalty \citep{ZouHastie2005}. It is now well-known that the elastic-net often outperforms the lasso ($\ell_1$ norm penalty) in prediction. \cite{WangEtAl2006} studied the elastic-net penalized SVM (DrSVM) and showed that the DrSVM performs better than the $\ell_1$ norm SVM.
Similarly,  we propose the elastic-net penalized DWD: 
\begin{equation}
\left(\hat\beta_0(\mathrm{enet}),\hat{\bs{\beta}}(\mathrm{enet})\right)=\argmin_{\beta_0, \bs{\beta}}\dfrac{1}{n}\sum_{i=1}^n V(y_i (\beta_0 + \mathbf{x}_i^T\bs{\beta})) + P_{\lambda_1, \lambda_2}(\bs{\beta}),
\label{eq:SHL4}
\end{equation}
where
$$
P_{\lambda_1, \lambda_2}(\bs{\beta})=\sum_{j=1}^p \left(\lambda_1|\beta_j|+\frac{\lambda_2}{2}\beta_j^2\right).
\label{eq:SHL3}
$$
The elastic-net penalized DWD classification rule is $\textrm{Sign}(\hat\beta_0(\mathrm{enet}) + \mathbf{x}^T\hat{\bs{\beta}}(\mathrm{enet}))$.
Both $\lambda_1$ and $\lambda_2$ are important tuning parameters for regularization. In practice, $\lambda_1$ and $\lambda_2$ are chosen from finite grids by validation or cross-validation. 



A further refinement of the elastic-net penalty is the adaptive elastic-net penalty \citep{ZouZhang2009}  where we replace the $\ell_1$ (lasso) penalty with the adaptive $\ell_1$ (lasso) penalty \citep{Zou2006}. The adaptive lasso penalty produces estimators with the oracle properties.
The adaptive elastic-net enjoys the benefits of elastic-net and adaptive lasso.  After fitting the elastic-net penalized DWD, we further consider the adaptive elastic-net penalized DWD:
\begin{equation}
\left(\hat\beta_0(\mathrm{aenet}),\hat{\bs{\beta}}(\mathrm{aenet})\right)=\argmin_{\beta_0, \bs{\beta}}\dfrac{1}{n}\sum_{i=1}^n V(y_i (\beta_0 + \mathbf{x}_i^T\bs{\beta})) + \sum_{j=1}^p \left(\lambda_1 \hat{\omega}_j|\beta_j|+\frac{\lambda_2}{2}\beta_j^2\right),
\label{eq:SHL11}
\end{equation}
and the adaptive weights are computed by
$$
\hat{\omega}_j=(|\hat{\beta}_j(\mathrm{enet})|+1/n)^{-1},
$$
where $\hat{\beta}_j(\mathrm{enet})$ is the solution of $\beta_j$ in \eqref{eq:SHL4}. 
The adaptive elastic-net penalized DWD classification rule is $\textrm{Sign}(\hat\beta_0(\mathrm{aenet}) + \mathbf{x}^T\hat{\bs{\beta}}(\mathrm{aenet}))$.

\section{Computation}
\label{sec:GCD}
The $\ell_2$ DWD was solved based on the second-order-cone programming; nevertheless, it is not trivial to generalize the algorithm to the $\ell_1$ DWD, and even more difficult to handle the elastic-net and the adaptive elastic-net penalties. In this section, we propose a completely different algorithm. We solve the solution paths of the sparse DWD by using the generalized coordinate descent (GCD) algorithm proposed by \cite{YangZou2013}. We introduce the GCD algorithm in section~\ref{sec:GCD1}, the implementation in section~\ref{sec:GCD3}, and the strict descent property in section \ref{sec:GCD2}. The same algorithm solves all the $\ell_1$, the elastic-net, and adaptive elastic-net penalized DWDs, while only the elastic-net is focused in the discussion for the sake of presentation. 

\subsection{Derivation of the algoithm}
\label{sec:GCD1}
Without loss of generality, we assume that the variables $\mathbf{x}_{j}$ are standardized: $\sum_{i=1}^n x_{ij}=0, \frac{1}{n}\sum_{i=1}^n x_{ij}^2=1$, for $j=1,\ldots,p$. We fix $\lambda_1$ and $\lambda_2$ and let $u_i = y_i(\tilde{\beta}_0 + \mathbf{x}_i^T \tilde{\bs\beta})$. We focus on $\beta_j$'s first. For each $\beta_j$, we define the coordinate-wise update function:
\begin{equation}
\begin{aligned}
F(\beta_j|\bs{\tilde{\beta}}, \tilde{\beta}_0)&=\frac{1}{n}\sum_{i=1}^n V\left(u_i+y_i x_{ij}(\beta_j-\tilde{\beta}_j)\right)+p_{\lambda_1,\lambda_2}(\beta_j).
\label{eq:alg1}
\end{aligned}
\end{equation}
Then the standard coordinate descent algorithm suggests cyclically updating 
\begin{equation}
\hat{\beta}_j = \argmin_{\beta_j}F(\beta_j|\tilde{\beta}_0, \tilde{\bs\beta})
\label{eq:alg50}
\end{equation}
for each $j=1, \ldots, p$. However, \eqref{eq:alg50} does not have a closed-form solution. The GCD algorithm solves this issue by adopting the MM principle \citep{HunterLange2004}. We approximate the $F$ function by a quadratic function
\begin{equation}
Q(\beta_j|\bs{\tilde{\beta}}, \tilde{\beta}_0)=\frac{\sum_{i=1}^n V(u_i)}{n}+\frac{\sum_{i=1}^n V'(u_i)y_i x_{ij}}{n}(\beta_j-\tilde{\beta}_j)+2(\beta_j-\tilde{\beta}_j)^2+p_{\lambda_1,\lambda_2}(\beta_j).
\label{eq:alg6}
\end{equation}
Then we update $\tilde{\beta_j}$ by 
$\tilde{\beta}_j^{\mathrm{new}}$, the closed-form minimizer of \eqref{eq:alg6}:
\begin{equation}
\tilde{\beta}_j^\text{new}=\dfrac{S \left(M\tilde{\beta}_j -\frac{1}{n}\sum_{i=1}^n V'(u_i)y_i x_{ij}, \lambda_1 \right)  }{4+\lambda_2},
\label{eq:alg7}
\end{equation}
where $S(z,r)=\text{sign}(z)(|z|-r)_+$ is the soft-thresholding operator \citep{DonohoJohnston1994} and $\omega_+ = \max(\omega, 0)$ is the positive part of $\omega$. 

With the intercept similarly updated, Algorithm~\ref{ALGO1} summarizes the details of the GCD algorithm. 
%

\begin{center}
\begin{algorithm}[ht]
\begin{enumerate}
\item Initialize $(\tilde{\beta}_{0},\tilde{\bs{\beta}})$.
\item Cyclic coordinate descent, for $j=1,2,\ldots,p$: 
\begin{enumerate}
\item
Compute $u_{i}=y_{i}(\tilde{\beta}_{0}+\mathbf{x}_{i}^{\intercal}\tilde{\bs{\beta}})$.
\item
Compute
$
\tilde{\beta}_{j}^{\text{new}}=
\frac{1}{4+\lambda_2} \cdot S\left(4\tilde{\beta}_{j}-\frac{1}{n}\sum_{i=1}^{n}V'(u_{i})y_{i}x_{ij},\lambda_{1}\right).
$
\item Set $\tilde{\beta}_j=\tilde{\beta}_j^{\text{new}}$.
\end{enumerate}
\item Update the intercept term: 
\begin{enumerate}
\item Compute $u_{i}=y_{i}(\tilde{\beta}_{0}+\mathbf{x}_{i}^T\tilde{\bs{\beta}})$.
\item Compute $\tilde{\beta}_{0}^{\text{new}}=\tilde{\beta}_{0}-{\sum_{i=1}^{n}V'(u_{i})y_{i}}/{(4n)}.$
\item Set $\tilde{\beta}_{0}=\tilde{\beta}_{0}^{\text{new}}$.
\end{enumerate}
\item Repeat steps 2-3 until convergence of $(\tilde{\beta}_{0},\tilde{\bs{\beta}})$.
\end{enumerate}
\caption{\textit{The GCD algorithm for the sparse DWD} \label{ALGO1}}
\end{algorithm}
\end{center}

\subsection{Implementation}
\label{sec:GCD3}
We have implemented Algorithm 1 in an R package \texttt{sdwd}. We exploit the warm-start, the strong rule, and the active set trick to increase the algorithm speeding. In our implementation, $\lambda_2$ is pre-chosen and we compute the solution path as $\lambda_1$ varies.

First, we adopt the warm-start to lead to a faster and more stable algorithm \citep{FriedmanEtAl2007}. We compute the solutions at a grid of $K$ decreasing $\lambda_1$ values, starting at the smallest $\lambda_1$ value such that $\boldsymbol{\tilde{\beta}}=0$. Denote these grid points by $\lambda_1^{[1]}, \ldots, \lambda_1^{[K]}$. With the warm-start trick, we can use the solution at $\lambda_1^{[k]}$ as the initial value (the warm-start) to compute the solution at $\lambda_1^{[k+1]}$.

Specifically, to find $\lambda_1^{[1]}$, we fit a model with a sufficiently large $\lambda_1$ and thus $\boldsymbol{\tilde{\beta}}=0$. Let $\hat{\beta}_0$ be the estimate of the intercept. By the KKT conditions, $\frac{1}{n} \max_j \left|\sum_{i=1}^n V'(\hat{\beta}_0) y_i x_{ij})\right| \le \lambda_1$, so we can choose
\begin{equation*}
\lambda_1^{[1]} = \frac{1}{n} \max_j \left|\sum_{i=1}^n V'(\hat{\beta}_0) y_i x_{ij})\right|.
\end{equation*}
Generally, we use $K=100$, and $\lambda_1^{[100]}=\epsilon \lambda_1^{[1]}$, where $\epsilon=10^{-4}$ when $n<p$ and $\epsilon=10^{-2}$ otherwise. All the other grid points are placed to uniformly distribute on a log scale.

Second, we follow the strong rule \citep{TibshiraniEtAl2010} to improve the computational speed. Suppose $\boldsymbol{\tilde{\beta}}^{[k]}$ and $\tilde\beta_0^{[k]}$ are the solutions at $\lambda_1^{[k]}$. After we solve $\boldsymbol{\tilde{\beta}}^{[k]}$ and $\tilde\beta_0^{[k]}$, the strong rule claims that any $j \in \{1, \ldots, p \}$ satisfying
\begin{equation}
\left|\frac{1}{n}\sum_{i=1}^n  V'(y_i(\hat\beta_0^{[k]} + x_i^T \boldsymbol{\hat{\beta}}^{[k]}))y_i x_{ij}\right| < 2 \lambda_1^{[k+1]} - \lambda_1^{[k]}
\label{eq:alg21}
\end{equation}
is likely to be inactive at $\lambda_1^{[k+1]}$, i.e., $\hat{\beta}_j^{[k+1]}=0$. Let $\mathcal{D}$ be the collection of $j$ which satisfies \eqref{eq:alg21}, and its compliment $\mathcal{D}^C=\{1, \ldots, p \}\backslash \mathcal{D}$. We call $\mathcal{D}^C$ the survival set. If the strong rule guesses correctly, the variables contained in $\mathcal{D}$ are discarded, and we only apply Algorithm~\ref{ALGO1} to repeat the coordinate descent in the survival set $\mathcal{D}^C$. After computing the solution $\hat\beta_0$ and $\boldsymbol{\hat{\beta}}$, we need to check whether some variables are incorrectly discarded. We check this by the KKT condition,
\begin{equation}
\left|\dfrac{1}{n}\sum_{i=1}^n  V'(y_i(\hat\beta_0 + x_i^T \boldsymbol{\hat{\beta}}))y_i x_{ij}\right|\le \lambda_1.
\label{eq:alg22}
\end{equation}
If no $j\in \mathcal{D}$ violates \eqref{eq:alg22}, $\hat\beta_0$ and $\boldsymbol{\hat{\beta}}$ are the solutions at $\lambda_1^{[k+1]}$. We rephrase them as $\tilde\beta_0^{[k+1]}$ and $\boldsymbol{\tilde{\beta}}^{[k+1]}$. Otherwise, any incorrectly discarded variable should be added to the survival set $\mathcal{D}^C$. We update $\mathcal{D}$ by $\mathcal{D}=\mathcal{D}/U$ where
\begin{equation*}
U = \left\{j: j\in \mathcal{D} \text{ and } \left|\dfrac{1}{n}\sum_{i=1}^n  V'(y_i(\hat\beta_0 + x_i^T \boldsymbol{\hat{\beta}}))y_i x_{ij} \right| > \lambda_1\right\}.
\end{equation*}
After each update of $\mathcal{D}$, some incorrectly discarded variables are added back to the survival set. 

Third, the active set is also used to boost the algorithm speed. After we apply Algorithm~\ref{ALGO1} on the survival set $\mathcal{D}^C$, we only apply the coordinate descent on a subset $S$ of $\mathcal{D}^C$ till convergence, where $S=\left\{j: j\in \mathcal{D}^C \text{ and } \beta_j \neq 0 \right\}$. Then another cycle of coordinate descent is run on $\mathcal{D}^C$ to investigate if the active set $S$ changes. We finish the algorithm if no changes in $S$; otherwise, we update the active set $S$ and repeat the process.

In Algorithm~\ref{ALGO1}, the margin $u_i$ can be updated conveniently: if $\beta_j$ is updated by $\beta_j^{\text{new}}$, we update $u_i$ by $u_i + y_i x_{ij} (\beta_j^{\text{new}}-\beta_j)$.

Last, the default convergence rule in \texttt{sdwd} is $4(\tilde{\beta}_j^{\mathrm{new}} - \tilde{\beta}_j)^2 < 10^{-8}$ for all $j=0,1,\ldots, p$. 

\subsection{The strict descent property of Algorithm 1}
\label{sec:GCD2}
\cite{YangZou2013} showed the GCD algorithm enjoys descent property. In this section, we also show the GCD algorithm has a stronger statement, the strict descent property, when the GCD is used to solve the sparse DWD. We first elaborate the following majorization result, whose proof is deferred in the appendix.
\begin{Lemma}
$F(\beta_j|\boldsymbol{\tilde{\beta}}, \tilde{\beta}_0)$ is the coordinate-wise update function defined in \eqref{eq:alg1}, and $Q(\beta_j|\boldsymbol{\tilde{\beta}}, \tilde{\beta}_0)$ is the surrogate function defined in \eqref{eq:alg6}.
We have \eqref{eq:alg11} and \eqref{eq:alg12}:

\begin{align}
F(\beta_j|\boldsymbol{\tilde{\beta}}, \tilde{\beta}_0)&=Q(\beta_j|\boldsymbol{\tilde{\beta}}, \tilde{\beta}_0), \ \text{if}\ \beta_j=\tilde{\beta}_j,
\label{eq:alg11}\\ 
F(\beta_j|\boldsymbol{\tilde{\beta}}, \tilde{\beta}_0)&< Q(\beta_j|\boldsymbol{\tilde{\beta}}, \tilde{\beta}_0), \ \text{if}\ \beta_j\neq\tilde{\beta}_j.
\label{eq:alg12}
\end{align}
\label{lemma:p5}
\end{Lemma}

\vspace*{-1cm}

Given $\tilde{\beta}_j^{\text{new}}=\argmin_{\beta_j} Q(\beta_j|\tilde{\beta}_0, \tilde{\boldsymbol\beta})$, and assuming $\tilde{\beta}_j^{\text{new}} \neq \tilde{\beta}_j$, \eqref{eq:alg11} and \eqref{eq:alg12} imply the strict descent property of the GCD algorithm: $F(\tilde{\beta}_j^{\text{new}}|\boldsymbol{\tilde{\beta}}, \tilde{\beta}_0) < F(\tilde{\beta}_j|\boldsymbol{\tilde{\beta}}, \tilde{\beta}_0)$. It is because $F(\tilde{\beta}_j^{\text{new}}|\boldsymbol{\tilde{\beta}}, \tilde{\beta}_0) < Q(\tilde{\beta}_j^{\text{new}}|\boldsymbol{\tilde{\beta}}, \tilde{\beta}_0) < Q(\tilde{\beta}_j|\boldsymbol{\tilde{\beta}}, \tilde{\beta}_0) = F(\tilde{\beta}_j|\boldsymbol{\tilde{\beta}}, \tilde{\beta}_0)$.
Note that the original GCD paper only showed $F(\tilde{\beta}_j^{\text{new}}|\boldsymbol{\tilde{\beta}}, \tilde{\beta}_0) \le F(\tilde{\beta}_j|\boldsymbol{\tilde{\beta}}, \tilde{\beta}_0)$.

The arguments above prove that the objective function $F$ strictly decreases after updating all variables in a cycle, unless the solution does not change after each update. If this is the case, the algorithm stops. We show that the algorithm must stop at the right answer. 
Assuming $\tilde{\beta}_j=\tilde{\beta}_j^{\text{new}}$ for all $j$, \eqref{eq:alg7} implies:
$$
\tilde{\beta}_j=\dfrac{S(4\tilde{\beta}_j -\frac{1}{n}\sum_{i=1}^n V'(u_i)y_i x_{ij}, \lambda_1)}{4+\lambda_2}.
\label{eq:alg13}
$$
A straightforward algebra can show that for all $j$,
$$
\begin{aligned}
&\dfrac{1}{n}\sum_{i=1}^n  V'(u_i)y_i x_{ij}+\lambda_1 \text{sign}(\beta_j)+\lambda_2 \beta_j
=0,\ &\text{if } \beta_j\neq0;\\
&\left|\dfrac{1}{n}\sum_{i=1}^n  V'(u_i)y_i x_{ij}\right|\le \lambda_1,\ &\text{if } \beta_j=0,
\end{aligned}
$$
which is exactly the KKT conditions of the original objective function \eqref{eq:SHL4}. In conclusion, if the objective function does not change after a cycle, the algorithm necessarily converges to the correct solution satisfying the KKT condition.

\section{Simulation}
The simulation in this section aims to support the following three points: (1) the sparse DWD has highly competitive prediction accuracy with the sparse SVM and the sparse logistic regression; (2) the adaptive elastic-net penalized DWD performs the best in variable selection; (3) for the prediction accuracy, no single method among the $\ell_1$, the elastic-net, and the adaptive elastic-net penalized DWDs dominate the others in all situations. 

In this section, the response variables of all the data are binary. The dimension $p$ of the variables $\mathbf{x}_i$ is always 3000. Within each example, our simulated data consist of a training set, an independent validation set, and an independent test set. The training set contains 50 observations: 25 of them are from the positive class and the other 25 from the negative class. Models are fitted on the training data only, and we use an independent validation set of 50 observations to select the tuning parameters: $\lambda_2$ is selected from $10^{-4}$, $10^{-3}$, $10^{-2}$, $0.1$, 1, 5, and 10; $\lambda_1$ is searched along the solution paths. We compared the prediction accuracy (in percentage) on another independent test data set of 20,000 observations.

We followed \cite{MarronEtAl2007} to generate the first two examples. In example 1, the positive class is a random sample from $N_p (\bs \mu_+, \bs I_p)$, where $\bs I_p$ is the $p$ by $p$ identity matrix and $\bs \mu_+$ has all zeros except for $2.2$ at the first dimension; the negative class is from $N_p (\bs \mu_-, \bs I_p)$ with $\bs \mu_- = - \bs\mu_+$. In example 2, $80\%$ of the data are generated from the same distributions as example 1; for the other $20\%$ of the data, the positive class is drawn from $N_p (\bs \mu_+, \bs I_p)$ and negative class $N_p (-\bs \mu_+, \bs I_p)$ where $\bs\mu_+ = (100, 500, 0, \ldots, 0)$. We obtained the other three examples following \cite{WangEtAl2006}. In example 3, the positive class has a normal distribution with mean $\bs \mu_+$ and covariance $\bs \Sigma=\bs I_{p \times p}$, where $\bs \mu_+$ has 0.7 in the first five covariates and 0 in others; the negative class has the same distribution except for a different mean $\bs \mu_- =-\bs \mu_+$. In example 4 and 5, we consider the cases where the relevant variables are correlated. Two classes have the same distributions except for the covariance,
\begin{align*}
\boldsymbol\Sigma= \left( \begin{array}{ccc}
\boldsymbol\Sigma_{5\times5}^\star & \boldsymbol{0}_{5 \times (p-5)}\\
\boldsymbol{0}_{(p-5)\times 5} & \boldsymbol{I}_{(p-5)\times (p-5)} \end{array} \right).
\end{align*}
In example 4, the diagonal elements of $\bs \Sigma^\star$ are 1 and the off-diagonal elements are all equal to $0.7$. In example 5, the $(i, j)$th element of $\bs\Sigma^\star$ equals $0.7^{|i-j|}$. 

We compared the sparse DWD with the sparse SVM and the sparse logistic regression. Both the DWD and the logistic regression use the $\ell_1$, the elastic-net and the adaptive elastic-net penalties. We used R packages \texttt{sdwd} and \texttt{gcdnet} \citep{YangZou2013} to compute the sparse DWDs and the sparse logistic regressions respectively. The $\ell_1$ and the elastic-net SVMs were solved by using the code from \cite{WangEtAl2006} which does not handle the adaptive elastic-net penalty. Table~\ref{tab:simu1} presents the prediction accuracy results. In the first two examples, the $\ell_1$ DWD and the $\ell_1$ logistic regression perform the best. We attribute this good performance to the only one nonzero variable in the data, despite $20\%$ of outliers in example 2. In example 3, 4, and 5, we increase the number of nonzero variables to five. For all models, the elastic-net and the adaptive elastic-net penalties have similar performance, and both of them dominate the $\ell_1$ penalties. The elastic-net DWD produces the least prediction error in example 4 and 5. Table 3 compares the variable selection. In all cases, the adaptive elastic-net penalties address all relevant variables with relatively few mistakes. The $\ell_1$ penalties share similar performance in the first two examples.

\begin{table}[t]
\caption{Comparisons of mis-classification percentage on 300 training data, 300 validation data, and 20,000 test data, based on 200 replicates. The numbers in parentheses are the standard errors. For each example, the methods with the best performance are marked by black boxes.} 
\resizebox{\textwidth}{!}{\begin{minipage}{1.2\textwidth}
\label{tab:simu1}
\ra{1.3}
\centering
\begin{tabular}{rrrrrrrrrrrrrrrrrrrr}
 \toprule
&&\multicolumn{3}{c}{DWD} &\phantom{} & \multicolumn{2}{c}{SVM} &\phantom{} & \multicolumn{3}{c}{logistic}&\\
\cmidrule{3-5}\cmidrule{7-8}\cmidrule{10-12}
 && $\ell_1$ & enet & aenet && $\ell_1$ & enet && $\ell_1$ & enet & aenet&\\
   \midrule
 & Example 1 & \framebox{\textbf{1.42}} & 1.47 & 1.44 &  & 1.46 & 1.50 &   & \framebox{\textbf{1.42}} & 1.46 & 1.44 &\\ 
&  {\footnotesize{Bayes: 
    1.39}} & (0.01) & (0.02) & (0.01) && (0.01) & (0.02) && (0.01) & (0.02) & (0.02)&\\ 
 & Example 2 & 1.14 & 1.15 & 1.13 && 1.16 & 1.16 && \framebox{\textbf{1.11}} & 1.14 & 1.15 &\\ 
  &{\footnotesize{Bayes: 
   1.11}} & (0.01) & (0.01) & (0.01) &  & (0.01)&(0.01)&& (0.01) & (0.01) & (0.02) &\\ 
     & Example 3 & 6.41 & 6.25 & 6.21 &  & 6.45 & \framebox{\textbf{6.15}} && 6.40 & 6.21 & 6.22 &\\ 
 & {\footnotesize{Bayes: 
    5.88}} & (0.03) & (0.03) & (0.03) &  & (0.04) & (0.03)&& (0.03) & (0.03) & (0.03) &\\ 
 & Example 4 & 22.05 & \framebox{\textbf{21.48}} & 21.54 &  & 22.03 & 21.56 &  & 22.00 & 21.54 & 21.64 &\\ 
 & {\footnotesize{Bayes: 
    21.10}} & (0.07) & (0.07) & (0.05) &  & (0.06) & (0.05)&   & (0.06) & (0.06) & (0.06) &\\ 
 & Example 5 & 18.91 & \framebox{\textbf{18.74}} & 18.75 &  & 18.84 & 18.78 &   & 18.81 & 18.80 & 18.77 &\\ 
 & {\footnotesize{Bayes: 
    18.03}} & (0.07) & (0.05) & (0.05) &  & (0.06) & (0.05)&& (0.06) & (0.05) & (0.05) &\\ 
   \bottomrule
   \end{tabular}
\end{minipage}}
\end{table}

\begin{table}[ht]
\caption{Comparisons of the variable selection. C is the number of selected nonzero variables, and IC is the number of zero variables incorrectly selected into the model. The results are the medians over 200 replicates. } 
\resizebox{\textwidth}{!}{\begin{minipage}{1.25\textwidth}
\label{tab:simu2}
\ra{1.3}
\centering
\begin{tabular}{@{}lrrrrrrrrrrrrrrrrrrrrrr}\toprule
& \multicolumn{6}{c}{DWD} &\phantom{a} & \multicolumn{4}{c}{SVM} &\phantom{a} & \multicolumn{6}{c}{logistic}\\
\cmidrule{2-7}\cmidrule{9-12}\cmidrule{14-19}
& \multicolumn{2}{c}{$\ell_1$} & \multicolumn{2}{c}{enet}& \multicolumn{2}{c}{anet} &&\multicolumn{2}{c}{$\ell_1$} & \multicolumn{2}{c}{enet}  
&& \multicolumn{2}{c}{$\ell_1$} & \multicolumn{2}{c}{enet} & \multicolumn{2}{c}{aenet} \\
 & C & IC  & C & IC & C & IC && C & IC & C & IC  && C & IC & C & IC & C & IC \\
  \midrule
Example 1 & 1 & 0 & 1 & 2 & 1 & 0 &  & 1 & 0 & 1 & 4  &  & 1 & 0 & 1 & 4.5 & 1 & 0 \\ 
Example 2 & 1 & 0 & 1 & 0 & 1 & 0 &  & 1 & 0 & 1 & 1  &  & 1 & 0 & 1 & 0 & 1 & 0 \\ 
Example 3 & 5 & 0 & 5 & 5 & 5 & 0 &  & 5 & 0 & 5 & 2.5 &  & 5 & 1 & 5 & 7 & 5 & 0 \\ 
Example 4 & 4 & 1 & 5 & 8.5 & 5 & 1.5 &  & 4 & 0 & 5 & 7  &  & 4 & 1 & 5 & 14 & 5 & 2 \\ 
Example 5 & 4 & 1 & 5 & 3.5 & 5 & 0 &  & 4 & 0 & 5 & 2  &  & 4 & 1 & 5 & 6.5 & 5 & 0 \\ 
 \bottomrule
\end{tabular}
 \end{minipage}}
\end{table}

\section{Real Data Examples}
In this section we analyze four benchmark data. The data Arcene was obtained from \cite{FrankAsuncion2010}, the breast cancer data from \cite{GrahamEtAl2010}, the LSVT data from \cite{TsanasEtAl2014}, and the prostate cancer was from \cite{SinghEtAl2002}. We randomly split each data with a ratio 1:1 into a training set and a test set. On the training set, we fit the sparse DWD with imposing the elastic-net and the adaptive elastic-net penalties. With the same tuning parameter candidates in the simulation, we used a five folder cross validation to find the best pair of $(\lambda_1, \lambda_2)$ incurring the least mis-classification rate. Then we investigated the prediction accuracy of the selected model on the test set. As comparisons, we considered the sparse SVM and the sparse logistic regression. Every method was trained and tuned in the same way as the sparse DWD. All numerical experiments were carried out on an Intel Core i7-3770 (3.40 GHz) processor.

In Table~\ref{tab:realdata}, we reported the average mis-classification percentage on the test set from 200 independent splits. We observe that the classifiers achieving the least error in these four datasets are the adaptive elastic-net logistic regression, the elastic-net SVM, the elastic-net and the adaptive elastic-net DWDs. We also find all the differences are not quite large. For the sparse DWD, we get the same message as \cite{MarronEtAl2007} concluded for the standard DWD: ``it very often is competitive with the best of the others and sometimes is better." We also notice that the computation of the sparse DWD is the fastest in almost all cases. The timing of the SVM is much longer than other methods. A possible explanation is that the SVM uses the non-differentiable hinge loss function which makes the GCD algorithm not suitable for solving the sparse SVM. So far, the best algorithm for the sparse SVM is a LARS type algorithm \cite{WangEtAl2006}, which is very different from the GCD algorithm for the sparse DWD and logistic regression. It has been observed that coordinate descent may be faster than the LARS algorithm for solving the lasso penalized least squares \citep{FriedmanEtAl2007}. 

\begin{table}[ht]
\caption{The mean mis-classification percentage and timings (in seconds) for four benchmark datasets. All the timings include the five-folder cross validation. The timings of adaptive elastic-net methods include computing the weights. The numbers in parentheses are the standard errors. For each data, the methods with the best prediction accuracy are marked by black boxes.} 
\resizebox{\textwidth}{!}{\begin{minipage}{1.25\textwidth}
\label{tab:realdata}
\ra{1.25}
\centering
\begin{tabular}{rrrllrllrllrllrll}
\toprule
& \multicolumn{2}{c}{Arcene} &\phantom{} & \multicolumn{2}{c}{Breast}&\phantom{}& 
 \multicolumn{2}{c}{LSVT}&\phantom{}& \multicolumn{2}{c}{Prostate}\\ 
& \multicolumn{2}{c}{\scriptsize{$n=100$, $p=10000$}} &\phantom{} & \multicolumn{2}{c}{\scriptsize{$n=42$, $p=22283$}}&\phantom{}
&\multicolumn{2}{c}{\scriptsize{$n=126$, $p=309$}}&\phantom{}& \multicolumn{2}{c}{\scriptsize{$n=102$, $p=6033$}}\\ 
\cmidrule{2-3} \cmidrule{5-6} \cmidrule{8-9}  \cmidrule{11-12} \cmidrule{14-15} 
& error  & time && error  & time && error & time && error & time
\\ 
  \midrule
enet DWD  & 34.43 & 123.41  && 26.50 & 58.40 && 16.01 & 8.28  && \framebox{\textbf{10.22}} & 28.18 \\ 
  & (0.56)  & (5.16)    && (1.00)  & (1.90)   && (0.34)  & (0.23) && (0.30)  & (0.95)\\ 
aenet DWD  & 34.60 & 200.19  && 26.86 & 116.12 && \framebox{\textbf{15.92}} & 13.72  && 10.26 & 39.25  \\ 
 & (0.57)  & (9.24)    && (1.00)  & (3.78)   && (0.34)  & (0.29)  && (0.26)  & (1.24) \\ 
enet logistic  & 34.16 & 211.18  && 24.67 & 145.35   && 16.96 & 10.73  && 10.65 & 102.19 \\ 
& (0.58)  & (3.40)    && (1.00)  & (0.74)   && (0.37)  & (0.18) && (0.29)  & (1.56)\\ 
aenet logistic  & \framebox{\textbf{34.15}} & 393.03  &&  25.12 & 290.31  &&16.93 & 17.02   &&  10.75 & 189.44 \\ 
 & (0.57)  & (6.52)    && (0.87)  & (1.47)   &&  (0.37)  & (0.29)  && (0.29)  & (2.84)\\ 
enet SVM  & 35.10 & 7410.09 && \framebox{\textbf{23.95}} & 567.43  && 16.27& 63.10 &&10.56 & 2508.94 \\ 
& (0.67)  & (1465.68) && (1.00)  & (15.19)  && (0.37)  & (0.77) && (0.36)  & (0.77) \\ 
   \bottomrule
\end{tabular}
 \end{minipage}}
\end{table}

\section{Discussion}
In this article, we have proposed the sparse DWD for high-dimensional classification and developed an efficient algorithm to compute its solution path. 
We have shown that the sparse DWD has competitive prediction performance with the sparse SVM and the sparse logistic regression and is often faster to compute with the help of our algorithm. Thus, the sparse DWD is a valuable addition to the toolbox for high-dimensional classification. 

The generalized DWD defined in \cite{HallEtAl2005} minimizes the $q$th power of the inverse margins. When $q=1$, it reduces to the usual DWD. For computation considerations, 
\cite{MarronEtAl2007} choose to fix $q=1$, because it leads to a second order cone programming problem. We have found that our algorithm can be readily used to solve the sparse generalized DWD with any positive $q$. 
In our numerical study we tried the generalized DWD with $q=0.5,1,2,5,100$ and also tried to use cross-validation to select a data-driven $q$ value. Our numeric results indicated that using different $q$ values does not lead to significant differences in performance. We opt to leave those results to the technical report version of this paper. 

\section*{Acknowledgments}
The authors thank the editor, the associate editor, and two referees for their helpful comments and suggestions. 
%
\section*{Appendix}
\paragraph*{Proof of Lemma~\ref{lemma:p5}} 
\eqref{eq:alg11} is trivial. To prove \eqref{eq:alg12}, it suffices to show for any $a \neq b \in \mathbb{R}$, 
\begin{equation}
V(a) < V(b) + V'(b)(a-b) + 2(a-b)^2.
\label{eq:app1}
\end{equation}
First, it is not hard to check that the first-order derivative $V'(\cdot)$ is Lipschitz continuous, i.e., for any $a \neq b$,
\begin{equation}
|V'(a)-V'(b)|<4|a-b|.
\label{eq:app2}
\end{equation}
Let $g(a)=2a^2-V(a)$, then \eqref{eq:app2} shows $g'(a)\equiv 4a-V'(a)$ is strictly increasing. Therefore $g(a)$ is a strictly convex function, and its first-order condition leads to \eqref{eq:app1} directly.
\end{document}